\documentclass{article}
\usepackage{ijcai25}

\usepackage{times}
\usepackage{soul}
\usepackage{url}
\usepackage[hidelinks]{hyperref}
\usepackage[utf8]{inputenc}
\usepackage[small]{caption}
\usepackage{graphicx}
\usepackage{amsmath}
\usepackage{amsthm}
\usepackage{booktabs}
\usepackage{algorithm}
\usepackage{algorithmic}
\usepackage[switch]{lineno}
\usepackage{amssymb}
\usepackage[table]{xcolor} 
\usepackage{colortbl}            
\usepackage{multirow}
\usepackage{subcaption}
\usepackage{pifont}

\title{ADC-GS: Anchor-Driven Deformable and Compressed Gaussian Splatting for Dynamic Scene Reconstruction}

\author{
He Huang$^1$\thanks{Contribute equally.}\and
Qi Yang$^2$\footnotemark[1]\and
Mufan Liu$^{1}$\and
Yiling Xu$^1$\thanks{Corresponding authors.}\and
Zhu Li$^2$\\
\affiliations
$^1$Shanghai Jiao Tong University\\
$^2$University of Missouri-Kansas City\\
\emails
\{huanghe0429, sudo\_evan, yl.xu\}@sjtu.edu.cn,
littlleempty@gmail.com, 
lizhu@umkc.edu%
}

\begin{document}

\maketitle

\begin{abstract}
Existing 4D Gaussian Splatting methods rely on per-Gaussian deformation from a canonical space to target frames, which overlooks redundancy among adjacent Gaussian primitives and results in suboptimal performance. To address this limitation, we propose Anchor-Driven Deformable and Compressed Gaussian Splatting (ADC-GS), a compact and efficient representation for dynamic scene reconstruction. Specifically, ADC-GS organizes Gaussian primitives into an anchor-based structure within the canonical space, enhanced by a temporal significance-based anchor refinement strategy. To reduce deformation redundancy, ADC-GS introduces a hierarchical coarse-to-fine pipeline that captures motions at varying granularities. Moreover, a rate-distortion optimization is adopted to achieve an optimal balance between bitrate consumption and representation fidelity. Experimental results demonstrate that ADC-GS outperforms the per-Gaussian deformation approaches in rendering speed by 300\%-800\% while achieving state-of-the-art storage efficiency without compromising rendering quality. The code is released at \textit{https://github.com/H-Huang774/ADC-GS.git}.
\end{abstract}

\section{Introduction}
Dynamic scene reconstruction from multi-view input videos has received significant attention due to its wide applications. Beyond methods based on 3D Gaussian Splatting (3DGS) \cite{3dgs}, 4D Gaussian Splatting (4DGS) has demonstrated substantial advances for dynamic scene reconstruction due to its impressive visual quality with ultra-fast training speed compared to neural radiance fields (NeRFs) based methods \cite{nerfs}. 

Recent research on 4DGS approaches has primarily focused on two categories. The first category employs 4D Gaussians to approximate the 4D volumes of scenes \cite{stg,4dgaussian,realtime4dgs} through temporal opacity and polynomial functions for each Gaussian. Another category focuses on temporally deforming a canonical space to target frames \cite{e-d3dgs,4d-gs,d3dgs,dn-4dgs}, utilizing a multilayer perceptron (MLP) with latent embeddings to predict changes in 3D Gaussian attributes over time.

\begin{figure}
    \centering
    \includegraphics[width=\linewidth]{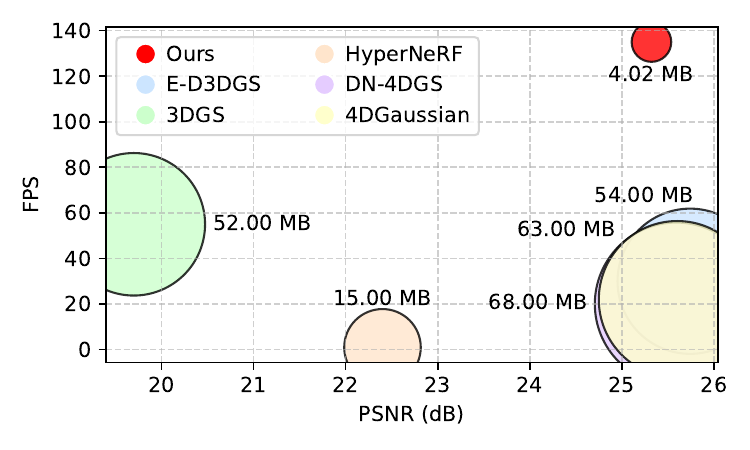}
    \vspace{-25pt}
    \caption{Comparison with concurrent dynamic scene reconstruction methods on the HyperNeRF dataset. Our method achieves the smallest storage size and the highest rendering speed while preserving excellent rendering quality.}
    \label{fig:psnr_vs_fps_size}
    \vspace{-15pt}
\end{figure}

Directly optimizing 4D Gaussians provides higher rendering speeds but requires a large number of Gaussians to model the entire sequence, resulting in both significant training time and substantial storage requirements. For example, \cite{realtime4dgs} often leads to over 9 hours of training time and 2GB of data for a 10-second video. Although deformation-based methods address the above issues and reduce training time, they still require a large number of Gaussians to achieve high-quality rendering. Consequently, substantial storage and bandwidth requirements emphasize the need for more compact 4DGS representations and advanced compression techniques, which is the primary focus of this work.

\begin{figure}[t]
    \centering
    \includegraphics[width=\linewidth]{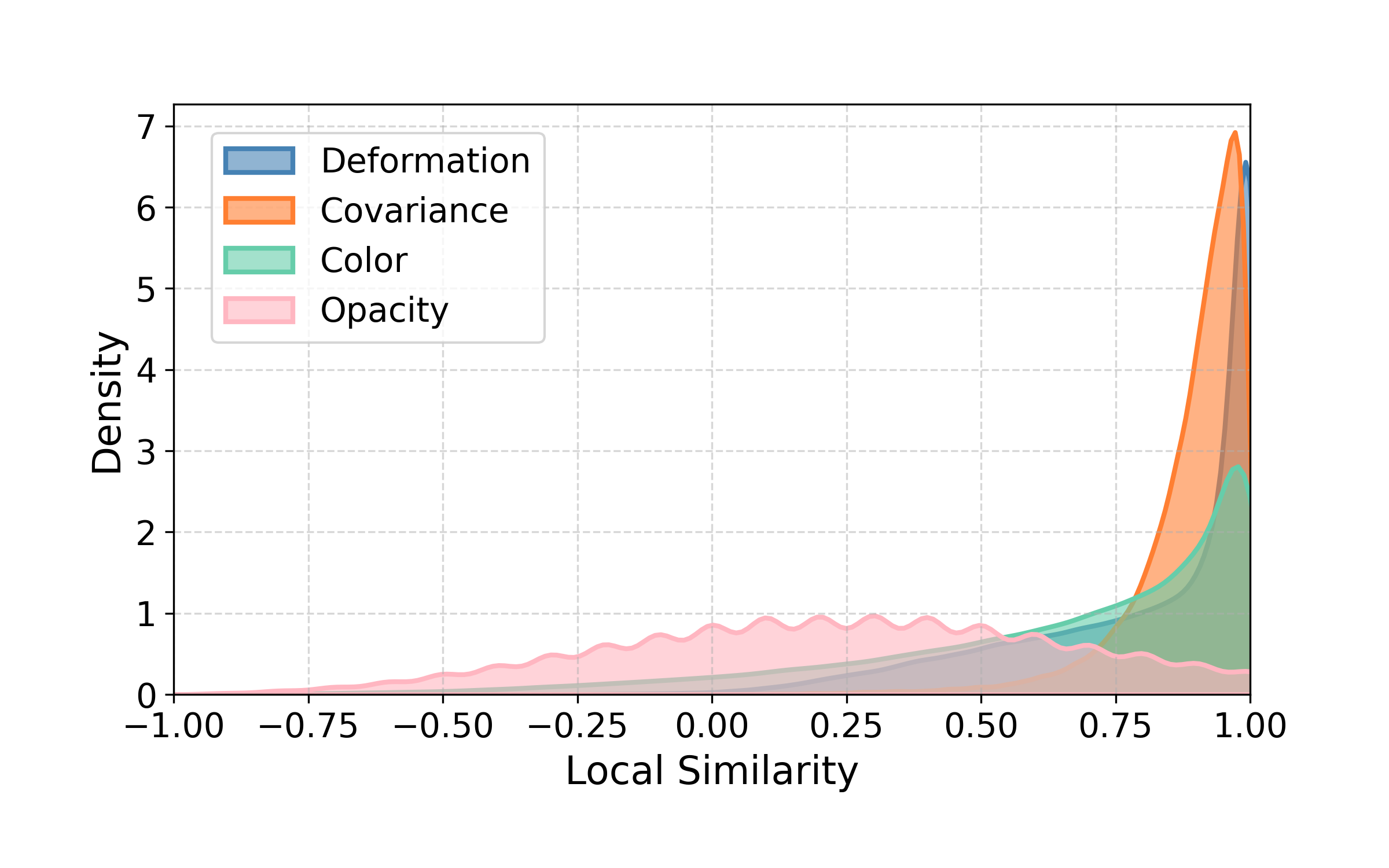}
    \vspace{-30pt}
    \caption{Illustration of local similarities of different features in \protect\cite{e-d3dgs}. The local similarity is measured by the average cosine distances between a Gaussian primitive and its 20 neighbors with minimal Euclidean distance.}
    \label{similarity}
    \vspace{-15pt}
\end{figure}

Some recent methods have developed effective technologies to reduce 4DGS storage. For example, 4DGaussian \cite{4d-gs} maps the 4D space onto six orthogonal planes as latent embeddings, effectively mitigating per-frame training redundancy and lowering storage costs. Since spherical harmonic coefficients dominate storage consumption, MEGA \cite{mega} decomposes them into a per-Gaussian DC color component and a lightweight AC color predictor, eliminating redundant coefficients to reduce storage requirements. 
However, these methods compress each Gaussian independently in the canonical space, neglecting the strong similarity between local attributes (excluding opacity), as shown in Figure \ref{similarity}. 
Furthermore, existing methods rely on per-Gaussian deformation for dynamic scene modeling, overlooking the consistent deformations among neighboring Gaussians since local scenes can often be approximated as rigid motion. These limitations hinder both the compactness of dynamic scene representations and rendering efficiency. 

Inspired by deformation-based dynamic scene reconstruction, this paper proposes \textbf{A}nchor-driven \textbf{D}eformable and \textbf{C}ompressed \textbf{G}aussian \textbf{S}platting (\textbf{ADC-GS}), a novel framework for efficient and compact dynamic scene representation. Specifically, we organize Gaussian primitives into a sparse set of anchors, facilitating streamlined representation and processing. These anchors, with \(K\) neural Gaussian primitives predicted from each by shared MLPs, collectively construct the canonical space.
To reconstruct dynamic scenes at any frame, we employ an anchor-based coarse-to-fine deformation strategy. In the coarse stage, the position, covariance, and color attributes of the anchors are deformed from the canonical space. Subsequently, the \(K\) Gaussian primitives are automatically updated based on the deformation of the associated anchor, notably reducing deformation redundancy. As the coarse stage focuses on capturing global deformation, a fine stage is introduced to dynamically refine the appearance of each primitive to recover fine-grained details. 
To further improve the compactness of the anchors, we implement a rate-distortion optimization scheme that incorporates a multi-dimension entropy model for accurate bitrate estimation and efficient compression.
Besides, we propose a novel anchor refinement method that leverages the accumulated gradients of each Gaussian primitive’s temporal significance to guide anchor growing, while using accumulated opacity to determine anchor pruning. This adaptive strategy robustly addresses both under-reconstruction and over-reconstruction issues in dynamic scenes, ensuring a balanced and efficient anchor representation. As shown in Figure \ref{fig:psnr_vs_fps_size}, ADC-GS achieves state-of-the-art (SOTA) storage size and rendering speed while preserving excellent rendering quality compared to prior works \cite{e-d3dgs,hypernerf,dn-4dgs,4d-gs,3dgs}.
Our contributions can be summarized as follows:
\begin{itemize}
    \item We propose anchor-driven deformable and compressed Gaussian Splatting (ADC-GS) for dynamic scene reconstruction. By leveraging compact anchors to efficiently model 4D scenes, our approach achieves an extraordinary storage reduction of up to 200\(\times\) over existing 4DGS methods.
    \item To accelerate rendering while preserving high reconstruction quality, we introduce a hierarchical coarse-to-fine deformation method. We also propose an adaptive anchor refinement strategy to address under-reconstruction and over-reconstruction issues.
    \item We develop a multi-dimension entropy model to enhance compactness through joint minimization of rendering distortion and bitrate consumption.
\end{itemize}

\begin{figure*}[t]
    \centering
    \includegraphics[width=\textwidth]{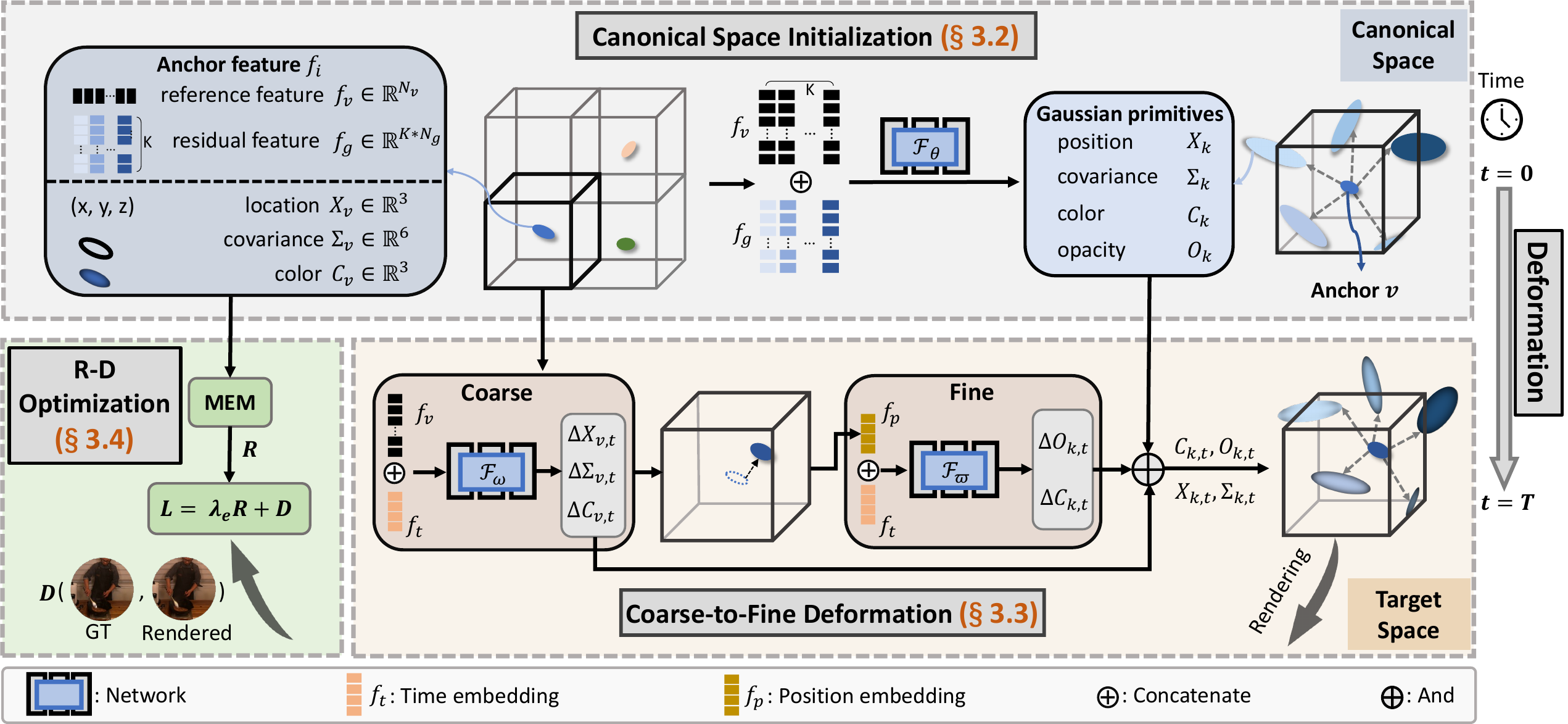}
    \vspace{-13pt}
    \caption{Overview of our ADC-GS framework. \textbf{Top:} ADC-GS organizes Gaussian primitives into a sparse set of anchors and compact residuals within canonical space. \textbf{Bottom right:} Gaussian primitives used for rendering are deformed from canonical space through a coarse-to-fine strategy based on anchors. \textbf{Bottom left:} Rendering distortion and estimated bitrates from the \textbf{MEM} are jointly minimized to balance rendering quality and storage efficiency.}
    \label{framework of ADC-GS}
    \vspace{-10pt}
\end{figure*}

\section{Related Work}
\subsection{Considering Spatial Relationships of Gaussian Primitives}
In 3D space, Scaffold-GS \cite{scaffold} and CompGS \cite{compgs} reconstruct scenes by synthesizing Gaussian primitives from anchors, exploiting the similarity of properties among neighboring Gaussians. HAC \cite{hac} utilizes a structured hash grid to take advantage of inherent consistencies among unorganized 3D Gaussians. In 4D scenes, some approaches \cite{d3dgs,gaussianprediction,ctymecd} introduce sparse control points combined with an MLP to model scene motion, based on the insight that motion can be effectively represented by a sparse set of basis. Inspired by the compact structures used in both 3D and 4D scenes, we design an anchor-driven deformable and compressed Gaussian splatting method to effectively eliminate intra-redundancy in both attributes and deformation among local Gaussian primitives.
\subsection{Deforming 3D Canonical Space}
Early research on deforming 3D canonical space to a target frame in both NeRF and 4DGS has explored various approaches. Nerfies \cite{nerfs} and HyperNeRF \cite{hypernerf} use per-frame trainable deformation rather than time-based conditions. D-NeRF \cite{dnerf} reconstructs dynamic scenes by deforming ray samples over time, utilizing a deformation network that takes 3D coordinates and timestamps as inputs. E-D3DGS \cite{e-d3dgs} defines the deformation as a function of Gaussian and temporal embeddings, decomposing the deformation into coarse and fine stages to model slow and fast movements, respectively. Other methods \cite{d3dgs,4d-gs,grid4d} learn implicit position mappings to deform primitives using latent embeddings, such as positional embeddings. However, these methods overlook the similarity in deformation among neighboring Gaussian primitives, resulting in suboptimal rendering speeds. To address this issue, we propose an anchor-based coarse-to-fine deformation method that accounts for deformation redundancy, achieving exceptional rendering speed while maintaining high reconstruction quality.
\subsection{3DGS Compression}
Recent studies on 3DGS compression can be categorized into processing-based and context-based methods. Processing-based methods include significance pruning \cite{lightgaussian}, scalar quantization \cite{compressed3dgs,fastgs}, codebooks \cite{compact3dgs}. Some point cloud compression techniques can also be adapted for Gaussian attributes due to their similar structure, such as graph-based signal transformation \cite{ggsc}, region-adaptive hierarchical transform \cite{hgsc}, and octree encoding \cite{octreecoding,ddpcc}. Context-based methods leverage various contextual information to remarkably compress Gaussian attributes, including hash-grid features \cite{hac}, hyperpriors \cite{compgs}, and hybrid Gaussian features \cite{feedforward}. Building on these advancements, we propose a multi-dimension entropy model for bitrate estimation and compression in this work, further eliminating redundancies within the anchors. By integrating the estimated bitrate into a rate-distortion optimization framework, our method effectively balances the trade-off between rendering quality and bitrate costs.

\section{Method}
\subsection{Overview}
As shown in Figure \ref{framework of ADC-GS}, ADC-GS begins with the initialization of a canonical space at \(t = 0\), characterized by a sparse set of anchors with associated Gaussian primitives. Each anchor is equipped with latent features and explicit Gaussian attributes, including position, covariance and color. The explicit Gaussian attributes serve as the foundation for the associated primitives, while the latent features are responsible for predicting the residuals and deformation of each primitive. 
To reconstruct any subsequent frame \(t = T\), we devise a coarse-to-fine strategy based on anchors. In the coarse stage, each anchor is deformed from the canonical space using lightweight multi-head MLPs, with the associated primitives subsequently updated based on the coarse deformation. In the fine stage, the opacity and color of each primitive are further refined using the deformed position of the anchors and current timestamp information.
By combining both coarse and fine deformations, all deformed attributes of Gaussian primitives are obtained to render the current frame using volume splatting \cite{3dgs}.
In the subsequent rate-distortion optimization, rendering distortion and estimated bitrates from the multi-dimension entropy model are jointly minimized to balance rendering quality and storage requirements.
Moreover, the temporal significance-based strategy is employed to prune and grow anchors for mitigating both under-reconstruction and over-reconstruction issues in dynamic scenes. 
\subsection{Canonical space initialization}
Unlike prior deformation-based approaches \cite{d3dgs,dn-4dgs} that directly use Gaussian primitives as the canonical space, we initialize our compact representation by employing a downsampled point cloud obtained via COLMAP \cite{colmap} to serve as anchors. These anchors are the foundation for local regions that improve overall efficiency by minimizing unnecessary duplication of information across the dynamic scene.

In ADC-GS, each anchor \(v\) is associated with \(K\) neural Gaussian primitives \(\{g_1,...,g_K\}\). The anchor consists of latent features \(\mathcal{A}_l = \{f_{v} \in \mathbb{R}^{N_{v}}, f_{g} \in \mathbb{R}^{K*N_g}\}\) and explicit Gaussian attributes \(\mathcal{A}_e = \{X_{v} \in \mathbb{R}^{3}, \Sigma_v \in \mathbb{R}^6, C_v \in \mathbb{R}^3\}\). Among the latent features, the reference feature \(f_{v}\) encapsulates the common characteristics between \(K\) associated primitives, while the residual feature \(f_{g}\) captures the variations specific to each primitive. The explicit Gaussian attributes serve as the foundation for the \(K\)  associated primitives.
To generate associated primitives from anchors, we predict their explicit attributes in residual forms, as: 
\begin{align}
\begin{split}
    (\Delta_{X_k}, \; \Delta_{\Sigma_k}, \; \Delta_{C_k}, \; {O}_k)
    &= \mathcal{F}_\theta(f_v, f_g);\\
    {X}_k = X_v + \Delta_{X_k}, \; {\Sigma}_k = \Sigma_v \Delta_{\Sigma_k},\; {C}_k &= C_v + \Delta_{C_k},
\end{split}
\end{align}
where \(\Delta_{X_k}, \Delta_{\Sigma_k}\) and \(\Delta_{C_k}\) are residuals of position, covariance, and color. \(\mathcal{F}_\theta\) represents the prediction network, which is modeled using a residual neural network. Since opacity \({O}_k\) exhibits no apparent spatial correlation as shown in Figure \ref{similarity}, it is directly predicted without residuals. 

Anchors and their associated primitives collectively define the canonical space that captures the global structure of the entire scene. Subsequently, the reconstruction of the scene in any frame \(t = T\) is performed by deforming this canonical space through a coarse-to-fine pipeline.
\subsection{Coarse-to-Fine Deformation}
Previous methods \cite{4d-gs,dn-4dgs} deform the geometric properties of each Gaussian primitive individually from the canonical space to the target frame \(T\) individually. However, this per-Gaussian deformation suffers from low efficiency due to the similar deformation among adjacent primitives, leading to slow rendering speeds. To mitigate the limitation, we propose an anchor-driven coarse-to-fine deformation strategy.
In the coarse stage, we deform the anchor explicit attributes \(\mathcal{A}_e\) by utilizing the concatenation of reference feature \(f_v\) and time embedding \(f_t\), formulated as:
\begin{align}
(\Delta X_{v,t},\; \Delta {\Sigma}_{v,t},\; \Delta C_{v,t}) = \mathcal{F}_\omega({f}_v, {f}_t),
\end{align}
where \(\mathcal{F}_\omega\) denotes the anchor deformation network, implemented as a tiny MLP. \(\Delta X_{v,t}, \Delta \Sigma_{v,t}, \Delta C_{v,t}\) denote the position, covariance and color deformation of the anchors. \(f_t\) encodes temporal information for different frames,
\begin{align}
    f_t = \mathcal{F}_s\left(\text{Interp}(Z), t \right]),
\end{align}
where \(Z \in \mathbb{R}^{256}\) is a learnable weight for interpolation, \(t\) denotes the temporal index and \(\mathcal{F}_s\) represents the grid sampling function to encode temporal variations. By deforming a small number of anchors, the associated \(K\) Gaussian primitives are automatically updated, substantially diminishing deformation redundancy to enhance rendering efficiency.

While the coarse stage captures the global changes of the associated Gaussian primitives, it lacks the capability to address finer details. Considering that opacity exhibits minimal local similarity and color significantly influences scene reconstruction \cite{e-d3dgs}, the fine stage dynamically refines these two attributes for each primitive, enhancing overall reconstruction quality.
This refinement integrates the position embedding \(f_p\) \cite{transformer} of the deformed anchors and the time embedding \(f_t\),
\begin{align}
    (\Delta O_{k,t}, \; \Delta C_{k,t}) = \mathcal{F}_\varpi({f}_p, {f}_t),
\end{align}
where \(\mathcal{F}_\varpi\) represents the fine-stage deformation network similar to \(\mathcal{F}_\omega\) and \(\Delta O_{k,t}, \Delta C_{k,t}\) denote the per-Gaussian refinements of opacity and color, respectively. Therefore, associated Gaussian primitives are derived as a combination of coarse and fine-stage deformations:
\begin{align}
\begin{split}
    {X}_{k,t} &= {X}_k + \Delta X_{v,t},\quad {\Sigma}_{k,t} = {\Sigma}_k\Delta \Sigma_{v,t}, \\
    {O}_{k,t} &= {O}_k + \Delta O_{k,t},\quad {C}_{k,t} = {C}_{k} + \Delta C_{v,t} + \Delta C_{k,t},
\end{split}
\end{align}
where \({X}_{k,t}\), \({\Sigma}_{k,t}\), \({O}_{k,t}\) and \({C}_{k,t}\) are the position, covariance, opacity and color attributes used for volume splatting in the target frame \(T\), respectively.

\begin{figure}
    \centering
    \includegraphics[width=\linewidth]{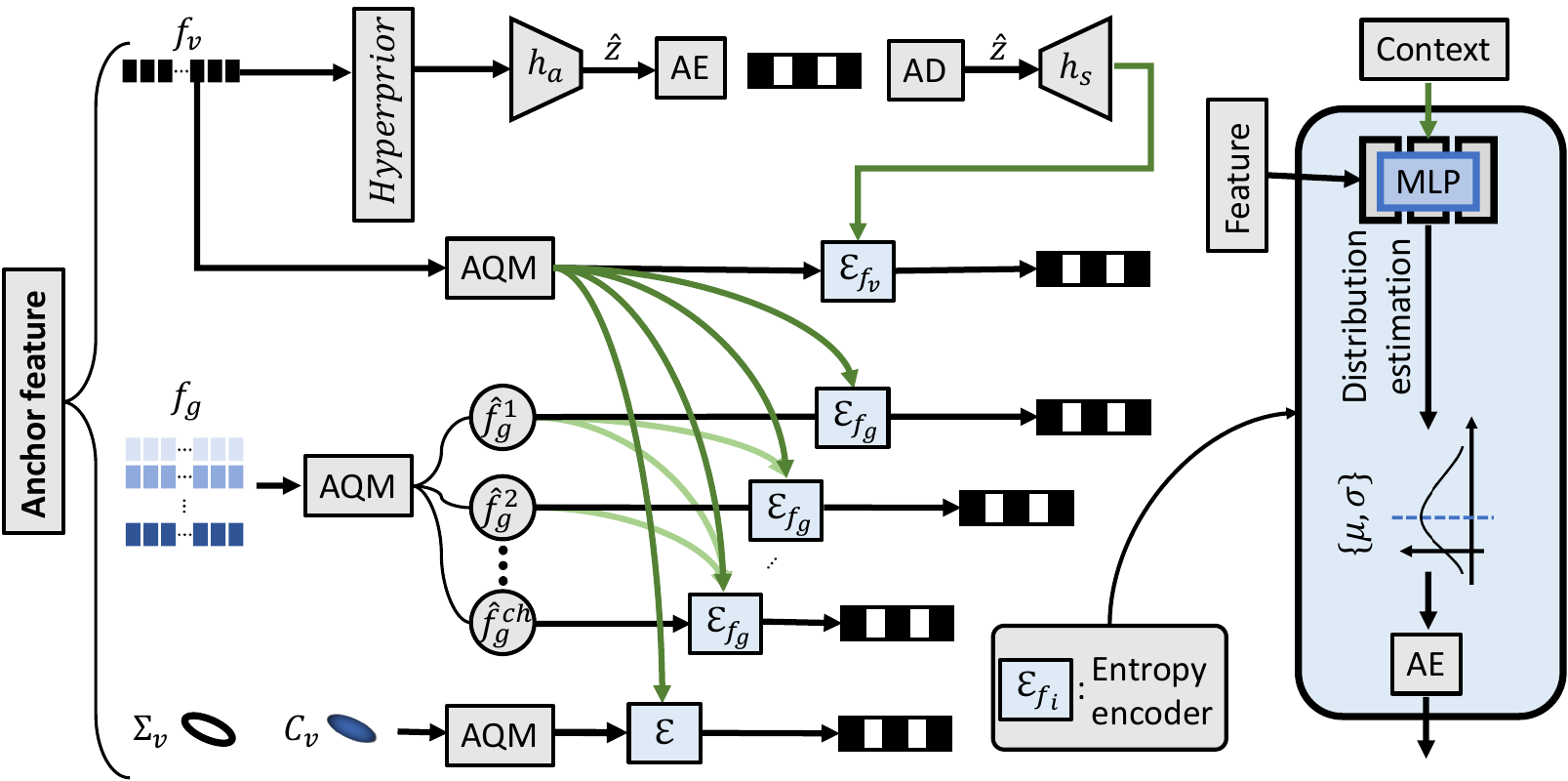}
    \caption{Illustration of the proposed MEM for accurate bitrates estimation. AQM refers to the adaptive quantization module.}
    \label{mem}
    \vspace{-10pt}
\end{figure}

\subsection{Rate-Distortion Optimization}
The rate-distortion optimization scheme is designed to achieve a more compact representation by jointly minimizing both bitrate consumption and rendering distortion. While the anchor locations \(X_v\) are compressed using G-PCC \cite{octreecoding}, the remaining four anchor features are modeled for bitrate estimation through a Multi-dimension Entropy Model (MEM), as demonstrated in Figure \ref{mem}. 
Specifically, scalar quantization is first applied to the feature \(f_i\), where \(f_i \in \{f_v, f_g, \Sigma_v, C_v\}\). However, conventional rounding essentially performs quantization with a fixed step size,  which is less flexible for features with different scales. To address this issue, we design an adaptive quantization module that better accommodates the diverse feature value scales as follows,
\begin{align}
    \hat{f}_i &= f_i + \mathcal{U}\left(-\frac{1}{2}, \frac{1}{2}\right) \times Q_i \times \left(1 + \tanh\left(\mathcal{F}_q(f_i)\right)\right),
\end{align}
where \(\mathcal{F}_q\) indicates MLP-based model to adjust the predefined quantization step size \(Q_i\). Note that \(Q_i\) varies for \(f_v, f_g, \Sigma_v\) and \(C_v\). Uniform noise \(\mathcal{U}(,)\) is injected during training to stimulate quantization loss in the testing stage as proposed in \cite{vae}.

Subsequently, the probability distribution of \(f_i\) is estimated to calculate the corresponding bitrate through the MLP-based context model \(\mathcal{E}_{f_i}\). The probability distribution \(p(\hat{f}_v)\) of reference feature is first parametrically formulated as a Gaussian distribution \(\mathcal{N}(\mu_{f_v}, \sigma_{f_v})\), where the parameters \(\{\mu_{f_v}, \sigma_{f_v}\}\) are predicted based on hyperpriors \cite{vae} extracted from \({f}_v\), 
\begin{align}
    p(\hat{f}_v) &= \mathcal{N}(\mu_{f_v}, \sigma_{f_v}), \quad \text{with} \quad \mu_{f_v}, \sigma_{f_v} = \mathcal{E}_{f_v}(\eta_{f_v}),
\end{align}
where \(\eta_{f_v}\) denotes the hyperpriors.
Furthermore, the decoded reference feature \(\hat{f}_v\) is used as contexts to model the probability distributions of covariance \(\hat{\Sigma}_v\), color \(\hat{C}_v\) and residual feature \(\hat{f}_g\).  The probability distribution of \(\hat{\Sigma}_v\) and \(\hat{C}_v\) is modeled similarly to \(p(\hat{f}_v)\).
The residual feature \(f_g\) constitutes the largest portion of the total bitstream, which is divided into \(M\) chunks. The decoded reference features \(\hat{f}_v\) and chunks are organized into a multi-dimensional context to guide the encoding of the remaining chunks. This strategy effectively reduces channel-wise redundancy and enhances compression efficiency,
\begin{align}
    \begin{split}
        p(\hat{f}_g^{ch}) &= \mathcal{N}(\mu_{f_g}, \sigma_{f_g}), \\
        \text{with} \quad 
        \mu_{f_g}, \sigma_{f_g} &= \mathcal{E}_{f_g}\left(\hat{f}_v \oplus \prod_{m=1}^{ch-1}\hat{f}_g^m\right),
    \end{split}
\end{align}
where \(ch\) means the index of current encoding chunks and \(\oplus\) denotes the channel-wise concatenation, effectively integrating the decoded context for improved compression efficiency.

Consequently, the total estimated bitrate consumption \(R\) of anchors is calculated as:
\begin{align}
\begin{split}
    R = &-\log_2 p(\hat{f}_v) - \log_2 p(\eta_{f_v}) - \\
    &\prod_{ch=1}^{M} \log_2 p(\hat{f}_{g}^{ch}) - \log_2 p(\hat{\Sigma}_v) - \log_2 p(\hat{C}_v).
\end{split}
\end{align}
Therefore, the rate-distortion optimization process is formulated as:
\begin{align}
    L_{loss} = (1 - \lambda_{ssim})L_1 + \lambda_{ssim}L_{ssim} + \lambda_{e}R.
\end{align}
where \(L_1\) and \(L_{ssim}\) represent the L1 loss and SSIM loss used in \cite{3dgs}. \(\lambda_{ssim}\) donates weighting coefficients for the SSIM loss and \(\lambda_{e}\) is the Lagrange multiplier to control the trade-off between rate and distortion.

\subsection{Anchors Refinement with Temporal Significance}
Anchors initialized from the sparse point cloud are often suboptimal, leading to under-reconstruction or over-reconstruction in certain areas. The existing strategies in 3DGS \cite{3dgs,scaffold} focus solely on static scenes and neglect the temporal significance of Gaussian primitives in dynamic frames, making them unsuitable for 4D scene reconstruction. To address this issue, we propose a temporal significance-aware anchor refinement method, including anchor growing and pruning. The importance of Gaussian primitives across frames is calculated by computing a weighted accumulation of their positional gradients,
\begin{align}
{\nabla}_k &= \frac{\sum^N \Psi(k,t) \|\nabla_{k,t}\|}{\sum^N \Psi(k,t)},
\end{align}
where \(\nabla_{k,t}\) is the 2D position gradient of a Gaussian primitive and \({\nabla}_k\) represents the accumulated gradients over \(N\) frames. \(\Psi(k,t)\) \cite{hgsc} is the rendering weight of the \(\alpha\)-blending in the target frame \(T\). Gaussian primitives with \({\nabla}_k\) values exceeding a predefined threshold are added as new anchors. These new anchors inherit latent features from their associated anchors while preserving their own Gaussian attributes. Conversely, anchors associated with Gaussian primitives whose opacity falls below a predefined threshold are pruned to avoid over-reconstruction.
\begin{figure}
    \centering
    \includegraphics[width=\linewidth]{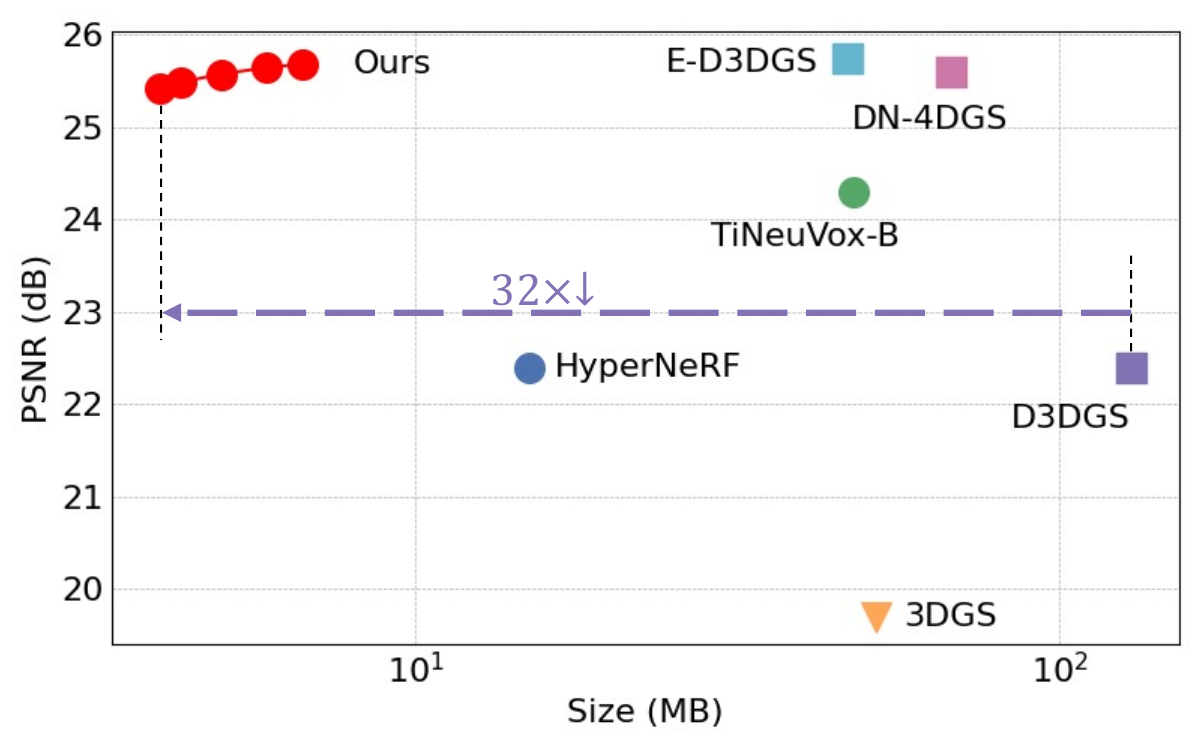}
    \vspace{-20pt}
    \caption{Rate-distortion curves of ADC-GS and comparison methods on HyperNeRF. We vary \(\lambda_e\) to achieve variable bitrates.}
    \label{rd-curve}
    \vspace{-10pt}
\end{figure}

\section{Experiments}
In this section, we present the implementation details of ADC-GS, followed by quantitative comparisons with SOTA approaches on two commonly used datasets. Finally, we provide in-depth analysis and ablation studies to evaluate the performance and effectiveness of the proposed ADC-GS.
\subsection{Experimental settings}
Our proposed ADC-GS framework is implemented based on Pytorch and trained on a single NVIDIA RTX 3090 GPU. For hyperparameter settings, the dimension of the reference feature \(N_v\) is set to 32, while the residual feature \(N_g\) is 16. Each anchor is associated with \(K = 10\) neural Gaussian primitives. The Lagrange multiplier \(\lambda_{e}\) is varied from \(e - 2\) to \(e - 4\) for different bitrates, and \(\lambda_{ssim}\) is fixed at 0.2. 
Quantization step sizes \(Q_i\) are set to 0.1, 0.1, 0.01 and 0.01 for \(f_v\), \(f_g\), \(\Sigma_v\) and \(C_v\), respectively. Meanwhile, \(f_g\) is divided into \(M = 4\) chunks. The prediction network \(\mathcal{F}_\theta\) and entropy estimation network \(\mathcal{E}_{f_i}\) are both implemented as 2-layer residual MLPs. Similarly, the deformation network \(\mathcal{F}_\omega\) and \(\mathcal{F}_{\varpi}\) are implemented as single 3-layer MLPs. The dimensions of time embedding \(f_t\) and position embedding \(f_p\) are 256 and 72, respectively.

\noindent\textbf{Dataset.} We evaluate our method on two multi-view video datasets: HyperNeRF \cite{hypernerf} and Neu3D \cite{dynerf}. HyperNeRF consists of videos captured using two phones rigidly mounted on a handheld rig, featuring four distinct scenes. Neu3D contains five distinct scenes, each comprising 20 multi-view videos. Data preprocessing follows the procedures outlined in \cite{4d-gs}.

\noindent\textbf{Baselines.} We compare our method against SOTA approaches from four categories. For NeRF-based methods, we include Nerfies \cite{nerfies}, HyperNeRF \cite{hypernerf}, TiNeuVox-B \cite{tineuvox}, DyNeRF \cite{dynerf}, NeRFPlayer \cite{nerfplayer}, HexPlane \cite{hexplane}, MixVoxels \cite{mixvoxel}, and HyperReel \cite{hyperreel}. For 3DGS-based compression methods, we compare with 3DGS \cite{3dgs} and CompGS \cite{compgs}. In addition, we use 3DGStream \cite{3dgstream}, Real-Time4DGS \cite{realtime4dgs}, and STG \cite{stg} as baselines for direct modeling of 4DGS methods. Finally, for deformation-based 4DGS methods, we benchmark against D3DGS \cite{d3dgs}, DN-4DGS \cite{dn-4dgs}, 4DGaussian \cite{4d-gs}, and E-D3DGS \cite{e-d3dgs}. 

\begin{table}[t]
\centering
\caption{Quantitative results on HyperNeRF. The best and 2nd best results are highlighted in \colorbox{red!30}{red} and \colorbox{yellow!50}{yellow}.}
\vspace{-5pt}
\label{hypernerf}
\resizebox{\linewidth}{!}{ 
\small
\begin{tabular}{lcccccc}
\toprule[2.5pt]
Model & \textbf{PSNR$\uparrow$} & \textbf{SSIM$\uparrow$} & \textbf{LPIPS$\downarrow$} & \textbf{FPS$\uparrow$} & \textbf{Size (MB)$\downarrow$} \\
\midrule
Nerfies & 22.20 & 0.803 & \cellcolor{yellow!50}0.170 & $<$ 1 & - \\
HyperNeRF & 22.40 & 0.814 & \cellcolor{red!30}0.153 & $<$ 1 & 15 \\
TiNeuVox-B & 24.30 & 0.836 & 0.393 & 1 & 48 \\
\midrule
3DGS & 19.70 & 0.680 & 0.383 & 55 & 52 \\
D3DGS & 22.40 & 0.612 & 0.275 & 22 & 129 \\
DN-4DGS & 25.59 & \cellcolor{red!30}0.861 & - & 20 & 68 \\
4DGaussian & 25.60 & \cellcolor{yellow!50}0.848 & 0.281 & 22 & 63 \\
E-D3DGS & \cellcolor{red!30}25.74 & 0.697 & 0.231 & 26 & 47 \\
\midrule
\multirow{3}{*}{\textbf{Ours}} & 25.42 & 0.777 & 0.315 & \cellcolor{red!30}135 & \cellcolor{red!30}4.02 \\
& 25.53 & 0.791 & 0.278 & \cellcolor{yellow!50}117 & \cellcolor{yellow!50}5.20 \\
& \cellcolor{yellow!50}25.68 & 0.825 & 0.252 & 101 & 6.67 \\
\bottomrule[2.5pt]
\end{tabular}
} 
\vspace{-0.2cm}
\end{table}

\noindent\textbf{Metrics.} We report the quantitative results of rendered images using peak-signal-to-noise ratio (PSNR), structural similarity index (SSIM) \cite{ssim}, perceptual quality measure LPIPS \cite{LPIPS} and rendering speed (FPS). Model size is also reported to assess compression efficiency. Three bitrate points are displayed for each dataset due to space limitations. Moreover, rate-distortion (RD) performance is included to ensure a fair comparison.

\subsection{Experimental Results}
\noindent\textbf{Quantitative results.} The quantitative results presented in Tables \ref{hypernerf}, \ref{neu3d} and Figure \ref{rd-curve} reveal consistent findings across all datasets: 1) our proposed ADC-GS achieves SOTA compression efficiency across all 4DGS methods, offering up to 32\(\times\) storage reduction on the HyperNeRF dataset compared to deformation-based 4DGS while maintaining excellent rendering quality. This improvement stems from organizing Gaussian primitives into anchor-driven residual forms, effectively mitigating intra-redundancy among adjacent primitives. Moreover, entropy loss leverages the MEM model to compress anchors, further enhancing storage efficiency. 2) All NeRF-based methods render at extremely low speeds due to their computationally intensive volumetric rendering process. Although direct modeling 4DGS methods achieve the highest FPS, they demand 204\(\times\) more storage compared to ADC-GS. Deformation-based methods demonstrate better convergence but still suffer from low FPS due to their per-Gaussian deformation strategy. In contrast, our proposed anchor-driven deformation method delivers rendering speeds up to 3\(\times\) faster than deformation-based methods, even surpassing some direct modeling 4DGS approaches. 3) ADC-GS only needs about 5MB of storage for both datasets, making it highly suitable for real-time transmission and practical applications.

\begin{table}[t]
\centering
\caption{Quantitative results on Neu3D. The best and 2nd best results are highlighted in \colorbox{red!30}{red} and \colorbox{yellow!50}{yellow}.}
\vspace{-5pt}
\label{neu3d}
\resizebox{\linewidth}{!}{ 
\begin{tabular}{lccccc}
\toprule[2.5pt]
Model & \textbf{PSNR$\uparrow$} & \textbf{SSIM$\uparrow$} & \textbf{LPIPS$\downarrow$} & \textbf{FPS$\uparrow$} & \textbf{Size (MB)$\downarrow$} \\
\midrule
DyNeRF & 29.58 & 0.980 & 0.083 & 0.015 & 28 \\
NeRFPlayer & 30.69 & 0.966 & 0.111 & 0.045 & - \\
HexPlane & 31.70 & \cellcolor{red!30}0.987 & 0.075 & 0.2 & 250 \\
Mixvoxels & 31.73 & - & 0.064 & 4.6 & 500 \\
HyperReel & 31.10 & 0.964 & 0.096 & 2.0 & 360 \\
\midrule
CompGS & 29.61 & 0.923 & 0.099 & 45 & 825 \\
3DGStream & 31.67 & - & - & \cellcolor{red!30}215 & 2340 \\
Real-Time4DGS & 32.01 & \cellcolor{yellow!50}0.986 & 0.055 & 114 & \>1000 \\
STG & \cellcolor{red!30}32.04 & 0.974 & 0.044 & 110 & 175 \\
DN-4DGS & \cellcolor{yellow!50}32.02 & 0.984 & \cellcolor{yellow!50}0.043 & 15 & 112 \\
4DGaussian & 31.72 & 0.984 & 0.049 & 34 & 38 \\
E-D3DGS & 31.20 & 0.974 & \cellcolor{red!30}0.030 & 42 & 40 \\
\midrule
\multirow{3}{*}{\textbf{Ours}} & 31.41 & 0.972 & 0.066 & \cellcolor{yellow!50}126 & \cellcolor{red!30}4.04 \\
 & 31.55 & 0.975 & 0.065 & 116 & \cellcolor{yellow!50}5.32 \\
 & 31.67 & 0.981 & 0.061 & 110 & 6.57 \\
\bottomrule[2.5pt]
\end{tabular}
}
\end{table}
\begin{figure}[t]
    \vspace{-10pt}
    \centering
    \includegraphics[width=\linewidth]{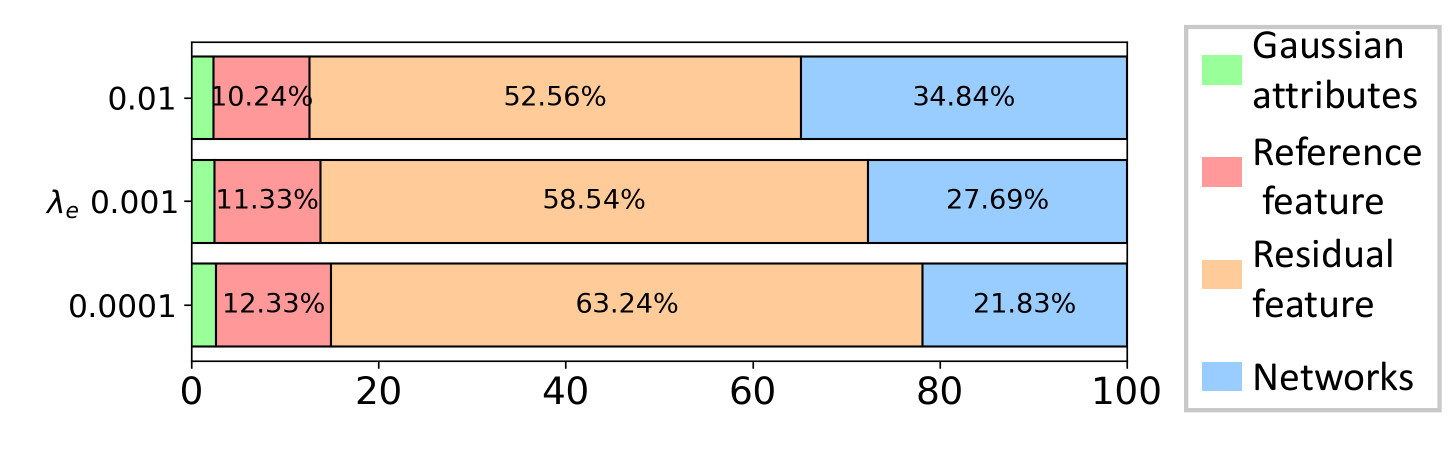}
    \vspace{-20pt}
    \caption{Bitstream analysis at multiple bitrate points on Neu3D.}
    \vspace{-15pt}
    \label{bitstream}
\end{figure}

\begin{figure*}[t]
    \centering
    \includegraphics[width=\textwidth]{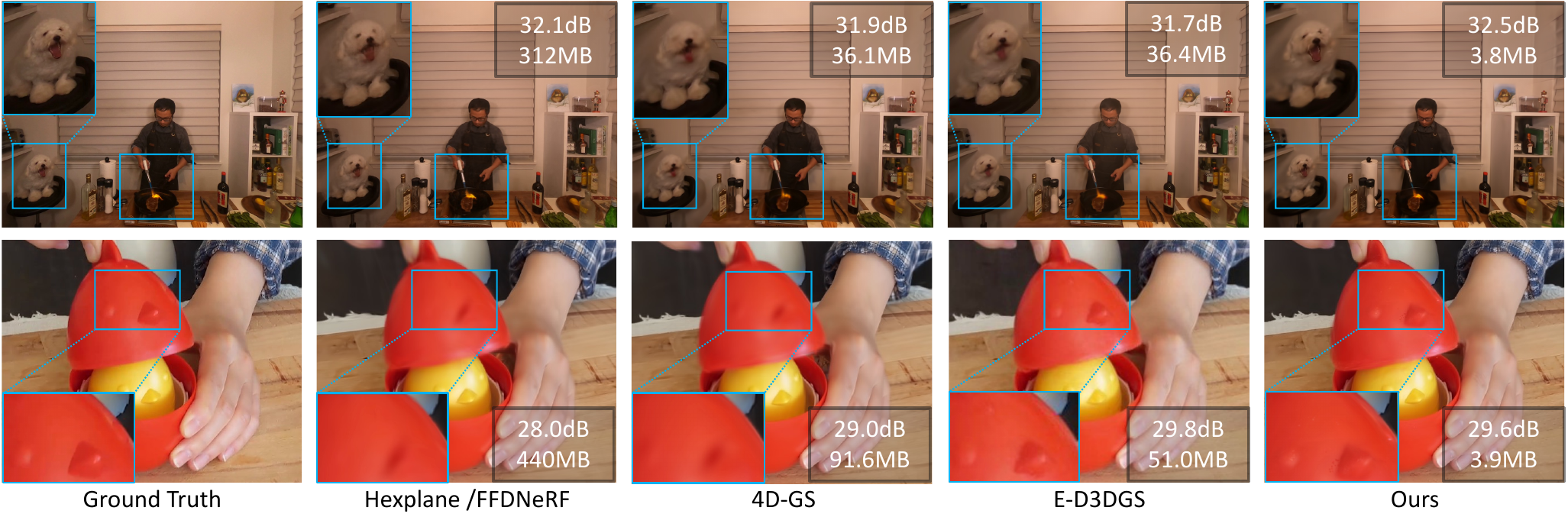}
    \vspace{-20pt}
    \caption{Qualitative quality comparisons of ``Flame Steak'' (Top) in Neu3D and ``Chicken'' (Bottom) in HyperNeRF dataset.}
    \vspace{-2pt}
    \label{visual}
\end{figure*}

 Figure \ref{visual} presents a qualitative comparison between ADC-GS and other approaches. The rendered images produced by ADC-GS exhibit unperceived distortion while saving lots of storage. Moreover, our method demonstrates clearer textures in certain datasets, attributed to the temporal significance-based anchor refinement that captures finer details, such as Flame Steak in Neu3D.

\noindent\textbf{Bitstream.} Our bitstream is summarized as four components: reference feature \(f_v\), residual feature \(f_g\), explicit Gaussian attributes (comprising covariance \(\Sigma_v\), color \(C_v\) and position \(X_v\)) and networks \(\mathcal{F}\). Among these, the networks are directly stored in 32 bits, \(X_v\) is compressed using G-PCC, and others are encoded using the AE entropy codec \cite{ae} with estimated probabilities from our MEM model. As demonstrated in Figure \ref{bitstream}, \(f_g\) constitutes the largest portion of the total bitstream, underscoring the importance of employing the MEM to effectively reduce channel-wise redundancy. The explicit Gaussian attributes account for a small portion of the bitstream, approximately 2\%. When analyzing the bit allocation of each component specifically, they are 0.60MB, 3.10MB, 0.15MB (0.10MB, 0.03MB, 0.02MB) and 1.47MB on Neu3D with \(\lambda_e = e - 3\), respectively. Notably, a smaller \(\lambda_{e}\) results in a larger bitstream allocation for features to enhance rendering quality.

\begin{table}[t]
    \centering
    \caption{Ablation study of different stages on ``HyperNeRF''. ``+'' indicates adding current module to the previous stage. ``Base'' is initialized canonical space.}
    \vspace{-0pt}
    \normalsize
    \label{ab_table}
    \resizebox{\columnwidth}{!}{  
    \renewcommand{\arraystretch}{1}  
    \begin{tabular}{l|cccc}
    \hline
         \textbf{Stages}  & \textbf{PSNR↑} & \textbf{SSIM↑} & \textbf{FPS↑} & \textbf{Size(MB)↓} \\ \hline
          Base &20.75 & 0.621 & 185 & 17.46 \\
            + Coarse Deformation & 24.83 & 0.676 & 119 & 17.77 \\ 
         + Fine Deformation & 25.03 & 0.689 & 117 & 18.11 \\ 
         + R-D Optimization & 25.00 & 0.687 & 112 & 4.49 \\ 
         + Anchors Refinement & 25.42 & 0.777 & 135 & 4.02 \\ \hline
    \end{tabular}
    }
\vspace{-20pt}
\end{table}

\subsection{Ablation Study}
\begin{figure}[t]
    \centering
    \includegraphics[width=\linewidth]{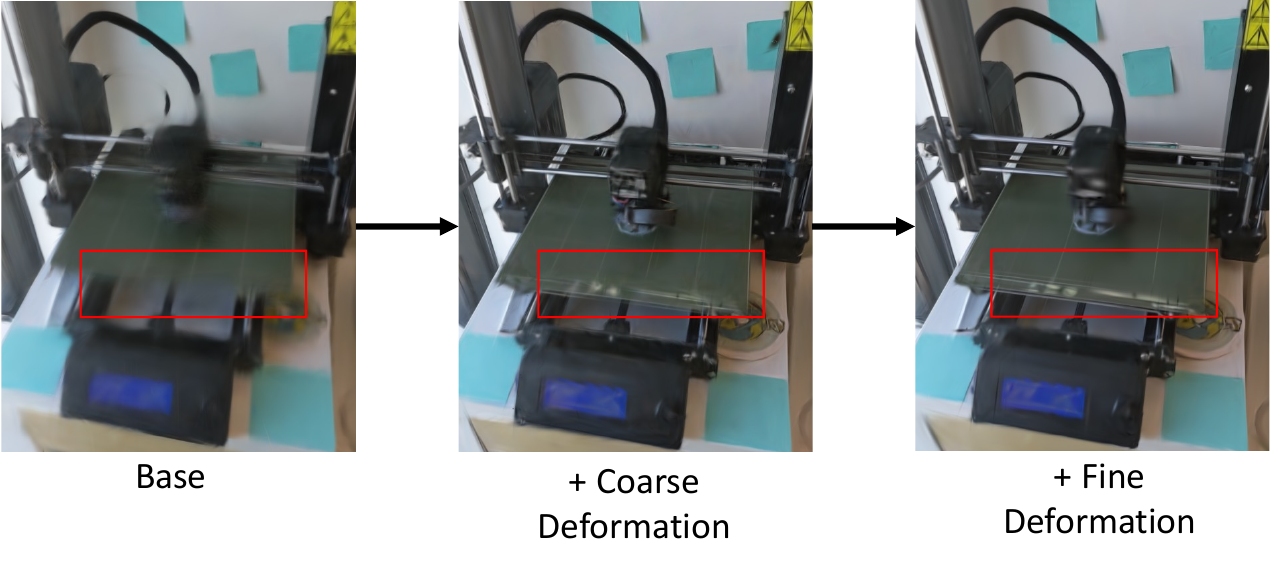}
    \vspace{-20pt}
    \caption{Ablation study of visual results for the coarse-to-fine deformation process.}
    \label{ab_printer}
    \vspace{-6pt}
\end{figure}

In this subsection, we evaluate the effectiveness of each component in our framework on the HyperNeRF dataset. As shown in Table \ref{ab_table} and Figure \ref{ab_printer}, the framework’s performance improves progressively with the addition of each module, demonstrating their individual and collective contributions. 

The base configuration includes only the initialized canonical space, achieving a significant size reduction compared to existing 4DGS methods \cite{4d-gs,dn-4dgs}. While this setup provides high rendering speed due to the absence of deformation, the reliance on canonical space to represent the entire scene results in blurred rendering, particularly in motion-heavy regions. The addition of the coarse deformation module significantly improves the ability to capture the global motion structure. Although the rendering speed decreases compared to the baseline, it remains considerably faster than the per-Gaussian deformation approach in \cite{e-d3dgs}, showcasing the efficiency of the anchor-driven strategy. The fine deformation module builds upon the coarse stage by refining appearance attributes, particularly in regions requiring higher detail. This refinement improves rendering quality by accurately modeling subtle temporal variations in color and opacity. The introduction of the rate-distortion optimization module leads to a remarkable reduction in bitstream size, from approximately 18 MB to 4.5 MB, without compromising rendering quality. This improvement can be attributed to the MEM module’s ability to learn compact anchor representations by effectively balancing bitrate consumption and reconstruction fidelity. Finally, the anchor refinement module further improves the representation by dynamically growing and pruning anchors to better capture fine details. This adaptive strategy ensures that the model maintains high-quality reconstruction in complex areas while optimizing the anchor distribution across the scene. 

Overall, each module contributes to achieving a balanced trade-off between compression efficiency, rendering speed, and dynamic scene reconstruction quality, resulting in a compact and effective representation of the scene. 

\section{Conclusion}
This paper introduces Anchor-Driven Deformable and Compressed Gaussian Splatting (ADC-GS), a novel method for dynamic scene reconstruction. By organizing Gaussian primitives in an anchor-based compact format within the canonical space, ADC-GS effectively minimizes the intra-redundancy between adjacent primitives. For dynamic scene modeling, we introduce a coarse-to-fine pipeline based on anchors instead of traditional per-Gaussian deformation, significantly reducing deformation redundancy. Meanwhile, a multi-dimension entropy model is employed to estimate the bitrate costs of anchors, serving as the factor for rate-distortion optimization. To further enhance details, we employ a temporal significance-based anchor refinement strategy. Extensive experiments demonstrate the effectiveness of ADC-GS and its technical components. Overall, our approach not only outperforms deformation-based 4DGS methods in rendering speed but also achieves SOTA storage efficiency without compromising rendering quality.
\section*{Acknowledgment}
This paper is supported in part by National Natural Science Foundation of China (62371290), National Key R\&D Program of China (2024YFB2907204), the Fundamental Research Funds for the Central Universities of China, and STCSM under Grant (22DZ2229005). The corresponding author is Yiling Xu(e-mail: yl.xu@sjtu.edu.cn).

\twocolumn[\newpage]

\bibliographystyle{named}
\bibliography{ijcai25}

\begin{thebibliography}{}

\bibitem[\protect\citeauthoryear{Attal \bgroup \em et al.\egroup }{2023}]{hyperreel}
Benjamin Attal, Jia-Bin Huang, Christian Richardt, Michael Zollhoefer, Johannes Kopf, Matthew O’Toole, and Changil Kim.
\newblock Hyperreel: High-fidelity 6-dof video with ray-conditioned sampling.
\newblock In {\em Proceedings of the IEEE/CVF Conference on Computer Vision and Pattern Recognition}, pages 16610--16620, 2023.

\bibitem[\protect\citeauthoryear{Bae \bgroup \em et al.\egroup }{2025}]{e-d3dgs}
Jeongmin Bae, Seoha Kim, Youngsik Yun, Hahyun Lee, Gun Bang, and Youngjung Uh.
\newblock Per-gaussian embedding-based deformation for deformable 3d gaussian splatting.
\newblock In {\em European Conference on Computer Vision}, pages 321--335. Springer, 2025.

\bibitem[\protect\citeauthoryear{Ball{\'e} \bgroup \em et al.\egroup }{2018}]{vae}
Johannes Ball{\'e}, David Minnen, Saurabh Singh, Sung~Jin Hwang, and Nick Johnston.
\newblock Variational image compression with a scale hyperprior.
\newblock {\em arXiv preprint arXiv:1802.01436}, 2018.

\bibitem[\protect\citeauthoryear{Cao and Johnson}{2023}]{hexplane}
Ang Cao and Justin Johnson.
\newblock Hexplane: A fast representation for dynamic scenes.
\newblock In {\em Proceedings of the IEEE/CVF Conference on Computer Vision and Pattern Recognition}, pages 130--141, 2023.

\bibitem[\protect\citeauthoryear{Chen \bgroup \em et al.\egroup }{2024a}]{ctymecd}
Tieyuan Chen, Huabin Liu, Tianyao He, Yihang Chen, Chaofan Gan, Xiao Ma, Cheng Zhong, Yang Zhang, Yingxue Wang, Hui Lin, et~al.
\newblock {MECD}: Unlocking multi-event causal discovery in video reasoning.
\newblock In {\em The Thirty-eighth Annual Conference on Neural Information Processing Systems}, 2024.

\bibitem[\protect\citeauthoryear{Chen \bgroup \em et al.\egroup }{2024b}]{feedforward}
Yihang Chen, Qianyi Wu, Mengyao Li, Weiyao Lin, Mehrtash Harandi, and Jianfei Cai.
\newblock Fast feedforward 3d gaussian splatting compression.
\newblock {\em arXiv preprint arXiv:2410.08017}, 2024.

\bibitem[\protect\citeauthoryear{Chen \bgroup \em et al.\egroup }{2025}]{hac}
Yihang Chen, Qianyi Wu, Weiyao Lin, Mehrtash Harandi, and Jianfei Cai.
\newblock Hac: Hash-grid assisted context for 3d gaussian splatting compression.
\newblock In {\em European Conference on Computer Vision}, pages 422--438. Springer, 2025.

\bibitem[\protect\citeauthoryear{Fan \bgroup \em et al.\egroup }{2022}]{ddpcc}
Tingyu Fan, Linyao Gao, Yiling Xu, Zhu Li, and Dong Wang.
\newblock D-dpcc: Deep dynamic point cloud compression via 3d motion prediction.
\newblock In Lud~De Raedt, editor, {\em Proceedings of the Thirty-First International Joint Conference on Artificial Intelligence, {IJCAI-22}}, pages 898--904. International Joint Conferences on Artificial Intelligence Organization, 7 2022.
\newblock Main Track.

\bibitem[\protect\citeauthoryear{Fan \bgroup \em et al.\egroup }{2023}]{lightgaussian}
Zhiwen Fan, Kevin Wang, Kairun Wen, Zehao Zhu, Dejia Xu, and Zhangyang Wang.
\newblock Lightgaussian: Unbounded 3d gaussian compression with 15x reduction and 200+ fps, 2023.

\bibitem[\protect\citeauthoryear{Fang \bgroup \em et al.\egroup }{2022}]{tineuvox}
Jiemin Fang, Taoran Yi, Xinggang Wang, Lingxi Xie, Xiaopeng Zhang, Wenyu Liu, Matthias Nie{\ss}ner, and Qi~Tian.
\newblock Fast dynamic radiance fields with time-aware neural voxels.
\newblock In {\em SIGGRAPH Asia 2022 Conference Papers}, pages 1--9, 2022.

\bibitem[\protect\citeauthoryear{Huang \bgroup \em et al.\egroup }{2024}]{hgsc}
He~Huang, Wenjie Huang, Qi~Yang, Yiling Xu, et~al.
\newblock A hierarchical compression technique for 3d gaussian splatting compression.
\newblock {\em arXiv preprint arXiv:2411.06976}, 2024.

\bibitem[\protect\citeauthoryear{Jiawei \bgroup \em et al.\egroup }{2024}]{grid4d}
Xu~Jiawei, Fan Zexin, Yang Jian, and Xie Jin.
\newblock {Grid4D}: {4D} decomposed hash encoding for high-fidelity dynamic scene rendering.
\newblock {\em The Thirty-eighth Annual Conference on Neural Information Processing Systems}, 2024.

\bibitem[\protect\citeauthoryear{Kerbl \bgroup \em et al.\egroup }{2023}]{3dgs}
Bernhard Kerbl, Georgios Kopanas, Thomas Leimk{\"u}hler, and George Drettakis.
\newblock 3d gaussian splatting for real-time radiance field rendering.
\newblock {\em ACM Trans. Graph.}, 42(4):139--1, 2023.

\bibitem[\protect\citeauthoryear{Lee \bgroup \em et al.\egroup }{2024}]{compact3dgs}
Joo~Chan Lee, Daniel Rho, Xiangyu Sun, Jong~Hwan Ko, and Eunbyung Park.
\newblock Compact 3d gaussian representation for radiance field.
\newblock In {\em Proceedings of the IEEE/CVF Conference on Computer Vision and Pattern Recognition}, pages 21719--21728, 2024.

\bibitem[\protect\citeauthoryear{Li \bgroup \em et al.\egroup }{2022}]{dynerf}
Tianye Li, Mira Slavcheva, Michael Zollhoefer, Simon Green, Christoph Lassner, Changil Kim, Tanner Schmidt, Steven Lovegrove, Michael Goesele, Richard Newcombe, et~al.
\newblock Neural 3d video synthesis from multi-view video.
\newblock In {\em Proceedings of the IEEE/CVF Conference on Computer Vision and Pattern Recognition}, pages 5521--5531, 2022.

\bibitem[\protect\citeauthoryear{Li \bgroup \em et al.\egroup }{2024}]{stg}
Zhan Li, Zhang Chen, Zhong Li, and Yi~Xu.
\newblock Spacetime gaussian feature splatting for real-time dynamic view synthesis.
\newblock In {\em Proceedings of the IEEE/CVF Conference on Computer Vision and Pattern Recognition}, pages 8508--8520, 2024.

\bibitem[\protect\citeauthoryear{Liu \bgroup \em et al.\egroup }{2024a}]{octreecoding}
Gexin Liu, Jiahao Zhu, Dandan Ding, and Zhan Ma.
\newblock Encoding auxiliary information to restore compressed point cloud geometry.
\newblock In Kate Larson, editor, {\em Proceedings of the Thirty-Third International Joint Conference on Artificial Intelligence, {IJCAI-24}}, pages 2189--2197. International Joint Conferences on Artificial Intelligence Organization, 8 2024.
\newblock Main Track.

\bibitem[\protect\citeauthoryear{Liu \bgroup \em et al.\egroup }{2024b}]{compgs}
Xiangrui Liu, Xinju Wu, Pingping Zhang, Shiqi Wang, Zhu Li, and Sam Kwong.
\newblock Compgs: Efficient 3d scene representation via compressed gaussian splatting.
\newblock In {\em Proceedings of the 32nd ACM International Conference on Multimedia}, pages 2936--2944, 2024.

\bibitem[\protect\citeauthoryear{Lu \bgroup \em et al.\egroup }{2024a}]{dn-4dgs}
Jiahao Lu, Jiacheng Deng, Ruijie Zhu, Yanzhe Liang, Wenfei Yang, Tianzhu Zhang, and Xu~Zhou.
\newblock Dn-4dgs: Denoised deformable network with temporal-spatial aggregation for dynamic scene rendering.
\newblock {\em arXiv preprint arXiv:2410.13607}, 2024.

\bibitem[\protect\citeauthoryear{Lu \bgroup \em et al.\egroup }{2024b}]{scaffold}
Tao Lu, Mulin Yu, Linning Xu, Yuanbo Xiangli, Limin Wang, Dahua Lin, and Bo~Dai.
\newblock Scaffold-gs: Structured 3d gaussians for view-adaptive rendering.
\newblock In {\em Proceedings of the IEEE/CVF Conference on Computer Vision and Pattern Recognition}, pages 20654--20664, 2024.

\bibitem[\protect\citeauthoryear{Ma \bgroup \em et al.\egroup }{2024}]{fastgs}
Yikun Ma, Dandan Zhan, and Zhi Jin.
\newblock Fastscene: Text-driven fast indoor 3d scene generation via panoramic gaussian splatting.
\newblock In Kate Larson, editor, {\em Proceedings of the Thirty-Third International Joint Conference on Artificial Intelligence, {IJCAI-24}}, pages 1173--1181. International Joint Conferences on Artificial Intelligence Organization, 8 2024.
\newblock Main Track.

\bibitem[\protect\citeauthoryear{Mildenhall \bgroup \em et al.\egroup }{2021}]{nerfs}
Ben Mildenhall, Pratul~P Srinivasan, Matthew Tancik, Jonathan~T Barron, Ravi Ramamoorthi, and Ren Ng.
\newblock Nerf: Representing scenes as neural radiance fields for view synthesis.
\newblock {\em Communications of the ACM}, 65(1):99--106, 2021.

\bibitem[\protect\citeauthoryear{Niedermayr \bgroup \em et al.\egroup }{2024}]{compressed3dgs}
Simon Niedermayr, Josef Stumpfegger, and R{\"u}diger Westermann.
\newblock Compressed 3d gaussian splatting for accelerated novel view synthesis.
\newblock In {\em Proceedings of the IEEE/CVF Conference on Computer Vision and Pattern Recognition}, pages 10349--10358, 2024.

\bibitem[\protect\citeauthoryear{Park \bgroup \em et al.\egroup }{2021a}]{nerfies}
Keunhong Park, Utkarsh Sinha, Jonathan~T Barron, Sofien Bouaziz, Dan~B Goldman, Steven~M Seitz, and Ricardo Martin-Brualla.
\newblock Nerfies: Deformable neural radiance fields.
\newblock In {\em Proceedings of the IEEE/CVF International Conference on Computer Vision}, pages 5865--5874, 2021.

\bibitem[\protect\citeauthoryear{Park \bgroup \em et al.\egroup }{2021b}]{hypernerf}
Keunhong Park, Utkarsh Sinha, Peter Hedman, Jonathan~T Barron, Sofien Bouaziz, Dan~B Goldman, Ricardo Martin-Brualla, and Steven~M Seitz.
\newblock Hypernerf: A higher-dimensional representation for topologically varying neural radiance fields.
\newblock {\em arXiv preprint arXiv:2106.13228}, 2021.

\bibitem[\protect\citeauthoryear{Pumarola \bgroup \em et al.\egroup }{2021}]{dnerf}
Albert Pumarola, Enric Corona, Gerard Pons-Moll, and Francesc Moreno-Noguer.
\newblock D-nerf: Neural radiance fields for dynamic scenes.
\newblock In {\em Proceedings of the IEEE/CVF Conference on Computer Vision and Pattern Recognition}, pages 10318--10327, 2021.

\bibitem[\protect\citeauthoryear{Snavely \bgroup \em et al.\egroup }{2006}]{colmap}
Noah Snavely, Steven~M. Seitz, and Richard Szeliski.
\newblock Photo tourism: exploring photo collections in 3d.
\newblock In {\em ACM SIGGRAPH 2006 Papers}, SIGGRAPH '06, page 835–846, New York, NY, USA, 2006. Association for Computing Machinery.

\bibitem[\protect\citeauthoryear{Song \bgroup \em et al.\egroup }{2023}]{nerfplayer}
Liangchen Song, Anpei Chen, Zhong Li, Zhang Chen, Lele Chen, Junsong Yuan, Yi~Xu, and Andreas Geiger.
\newblock Nerfplayer: A streamable dynamic scene representation with decomposed neural radiance fields.
\newblock {\em IEEE Transactions on Visualization and Computer Graphics}, 29(5):2732--2742, 2023.

\bibitem[\protect\citeauthoryear{Sun \bgroup \em et al.\egroup }{2024}]{3dgstream}
Jiakai Sun, Han Jiao, Guangyuan Li, Zhanjie Zhang, Lei Zhao, and Wei Xing.
\newblock 3dgstream: On-the-fly training of 3d gaussians for efficient streaming of photo-realistic free-viewpoint videos.
\newblock In {\em Proceedings of the IEEE/CVF Conference on Computer Vision and Pattern Recognition}, pages 20675--20685, 2024.

\bibitem[\protect\citeauthoryear{Vaswani}{2017}]{transformer}
A~Vaswani.
\newblock Attention is all you need.
\newblock {\em Advances in Neural Information Processing Systems}, 2017.

\bibitem[\protect\citeauthoryear{Wang \bgroup \em et al.\egroup }{2004}]{ssim}
Zhou Wang, A.C. Bovik, H.R. Sheikh, and E.P. Simoncelli.
\newblock Image quality assessment: from error visibility to structural similarity.
\newblock {\em IEEE Transactions on Image Processing}, 13(4):600--612, 2004.

\bibitem[\protect\citeauthoryear{Wang \bgroup \em et al.\egroup }{2023}]{mixvoxel}
Feng Wang, Sinan Tan, Xinghang Li, Zeyue Tian, Yafei Song, and Huaping Liu.
\newblock Mixed neural voxels for fast multi-view video synthesis.
\newblock In {\em Proceedings of the IEEE/CVF International Conference on Computer Vision}, pages 19706--19716, 2023.

\bibitem[\protect\citeauthoryear{Witten \bgroup \em et al.\egroup }{1987}]{ae}
Ian~H Witten, Radford~M Neal, and John~G Cleary.
\newblock Arithmetic coding for data compression.
\newblock {\em Communications of the ACM}, 30(6):520--540, 1987.

\bibitem[\protect\citeauthoryear{Wu \bgroup \em et al.\egroup }{2024}]{4d-gs}
Guanjun Wu, Taoran Yi, Jiemin Fang, Lingxi Xie, Xiaopeng Zhang, Wei Wei, Wenyu Liu, Qi~Tian, and Xinggang Wang.
\newblock 4d gaussian splatting for real-time dynamic scene rendering.
\newblock In {\em Proceedings of the IEEE/CVF Conference on Computer Vision and Pattern Recognition}, pages 20310--20320, 2024.

\bibitem[\protect\citeauthoryear{Yan \bgroup \em et al.\egroup }{2024}]{realtime4dgs}
Jinbo Yan, Rui Peng, Luyang Tang, and Ronggang Wang.
\newblock 4d gaussian splatting with scale-aware residual field and adaptive optimization for real-time rendering of temporally complex dynamic scenes.
\newblock In {\em Proceedings of the 32nd ACM International Conference on Multimedia}, pages 7871--7880, 2024.

\bibitem[\protect\citeauthoryear{Yang \bgroup \em et al.\egroup }{2024a}]{ggsc}
Qi~Yang, Kaifa Yang, Yuke Xing, Yiling Xu, and Zhu Li.
\newblock A benchmark for gaussian splatting compression and quality assessment study.
\newblock In {\em Proceedings of the 6th ACM International Conference on Multimedia in Asia}, pages 1--8, 2024.

\bibitem[\protect\citeauthoryear{Yang \bgroup \em et al.\egroup }{2024b}]{4dgaussian}
Zeyu Yang, Hongye Yang, Zijie Pan, and Li~Zhang.
\newblock Real-time photorealistic dynamic scene representation and rendering with 4d gaussian splatting.
\newblock In {\em International Conference on Learning Representations (ICLR)}, 2024.

\bibitem[\protect\citeauthoryear{Yang \bgroup \em et al.\egroup }{2024c}]{d3dgs}
Ziyi Yang, Xinyu Gao, Wen Zhou, Shaohui Jiao, Yuqing Zhang, and Xiaogang Jin.
\newblock Deformable 3d gaussians for high-fidelity monocular dynamic scene reconstruction.
\newblock In {\em Proceedings of the IEEE/CVF Conference on Computer Vision and Pattern Recognition}, pages 20331--20341, 2024.

\bibitem[\protect\citeauthoryear{Zhang \bgroup \em et al.\egroup }{2018}]{LPIPS}
Richard Zhang, Phillip Isola, Alexei~A. Efros, Eli Shechtman, and Oliver Wang.
\newblock The unreasonable effectiveness of deep features as a perceptual metric.
\newblock In {\em Proceedings of the IEEE Conference on Computer Vision and Pattern Recognition (CVPR)}, pages 586--595, 2018.

\bibitem[\protect\citeauthoryear{Zhang \bgroup \em et al.\egroup }{2024}]{mega}
Xinjie Zhang, Zhening Liu, Yifan Zhang, Xingtong Ge, Dailan He, Tongda Xu, Yan Wang, Zehong Lin, Shuicheng Yan, and Jun Zhang.
\newblock Mega: Memory-efficient 4d gaussian splatting for dynamic scenes.
\newblock {\em arXiv preprint arXiv:2410.13613}, 2024.

\bibitem[\protect\citeauthoryear{Zhao \bgroup \em et al.\egroup }{2024}]{gaussianprediction}
Boming Zhao, Yuan Li, Ziyu Sun, Lin Zeng, Yujun Shen, Rui Ma, Yinda Zhang, Hujun Bao, and Zhaopeng Cui.
\newblock Gaussianprediction: Dynamic 3d gaussian prediction for motion extrapolation and free view synthesis.
\newblock In {\em ACM SIGGRAPH 2024 Conference Papers}, pages 1--12, 2024.

\end{thebibliography}


\clearpage           
\appendix            

\maketitle
\section{Training Process}
\begin{figure}[h]
    \centering
    \includegraphics[width=\linewidth]{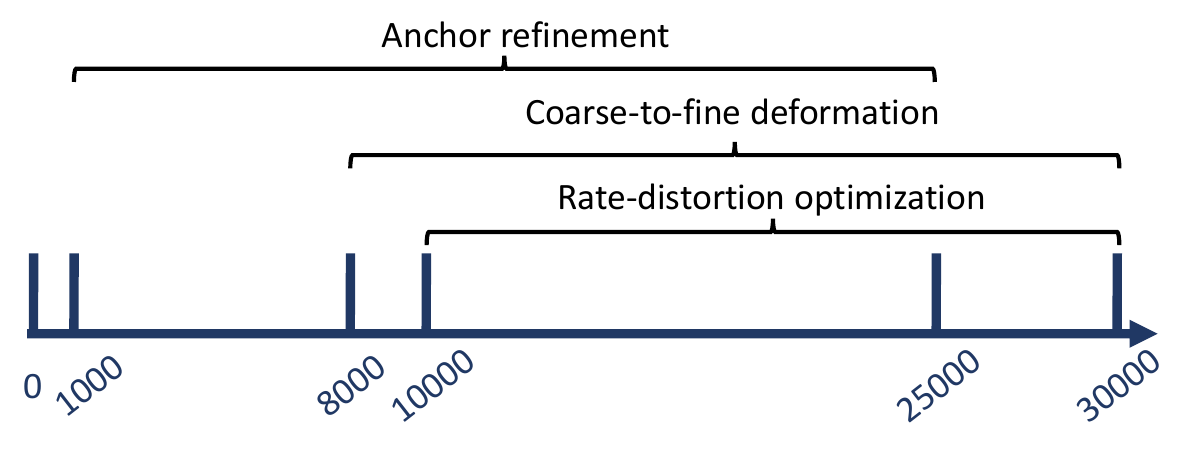}
    \caption{Detailed training process of our ADC-GS model.}
    \label{milestones}
\end{figure}

We present a comprehensive overview of the training process for our ADC-GS framework, as depicted in Figure \ref{milestones}.

\noindent\textbf{From iteration 1000 to 25000,} anchor refinement is employed to enhance the level of detail. It is important to note that the canonical space is initialized at the beginning, with anchors downsampled from a sparse set of points, which typically exhibit low quality. To address issues of under- and over-reconstruction, we apply anchor growing and pruning methods. Unlike the traditional 3DGS growing strategy, we incorporate temporal significance to better capture the importance of each Gaussian primitive. The temporal significance \(\Psi(k,t)\) is the 2D rendering weight, formulated as:
\begin{align}
    \Psi(k,t) = \sum_{p \in \mathcal{P}} \alpha_{i,t} \prod_{j=1}^{i-1} (1 - \alpha_{j,t}),
\end{align}
where \(\mathcal{P}\) represents the set of pixels that are overlapped by the projection of the Gaussian, while \(i\) denotes the set of sorted Gaussians along the ray which is determined by the \(\alpha-\)blending. The term \(\alpha\) measures the opacity of each Gaussian after 2D projection. 

Following refinement, the anchor positions are subjected to voxelization and compression using the G-PCC codec, after which they remain fixed and unoptimized.

\noindent\textbf{After iteration 8000,} we apply coarse-to-fine deformation for dynamic scene modeling. Optimization of the canonical space during the preceding iterations accelerates the convergence of our model.

\noindent\textbf{After iteration 10000,} we introduce rate-distortion optimization to improve the trade-off between bitrate consumption and rendering quality, while also making the anchors more compact.

\section{Ablation study on the proportion of Gaussian primitives.}

\begin{table}[ht]
\centering
\caption{Ablation studies on the proportion of Gaussian primitives on HyperNeRF.}
\label{tb:hypernerf}
\begin{tabular}{c|ccccc}
\hline
\textbf{K} & \textbf{PSNR} & \textbf{SSIM} & \textbf{LPIPS} &\textbf{FPS} & \textbf{Size (MB)} \\ \hline
5           & 24.39             & 0.756         & 0.337          & 172        & 3.73  \\ 
10          & 25.42             & 0.777         & 0.315          & 135         & 4.02   \\ 
15          & 25.28             & 0.775         & 0.328          & 81          & 8.28   \\ \hline
\end{tabular}
\end{table}

\begin{table}[ht]
\centering
\caption{Ablation studies on the proportion of Gaussian primitives on Neu3D.}
\label{tb:neu3d}
\begin{tabular}{c|ccccc}
\hline
\textbf{K} & \textbf{PSNR} & \textbf{SSIM} & \textbf{LPIPS} & \textbf{FPS} & \textbf{Size (MB)} \\ \hline
5           & 30.14         & 0.946         & 0.071          & 161       & 3.65   \\ 
10          & 31.41         & 0.972         & 0.066          & 126          & 4.04   \\ 
15          & 31.55         & 0.979         & 0.059          & 76        & 9.32   \\ \hline
\end{tabular}
\end{table}
As shown in Tables \ref{tb:hypernerf} and \ref{tb:neu3d}, we adjust the proportion of coupled primitives by varying the number of Gaussian primitives \(K\) associated with each anchor. Increasing \(K\) from 5 to 10 results in a significant improvement in PSNR. However, increasing \(K\) from 10 to 15 does not lead to a noticeable gain in PSNR, while doubling the size. Therefore, the \(K = 10\) setting provides the best balance, which prompts us to set \(K\) to 10 in our experiments.

\section{Complexity Analysis}
\noindent\textbf{Training Time.} The integration of additional components in ADC-GS increases the overall training time, approximately doubling that of other methods. Specifically, E-D3DGS requires an average of 35 minutes to train on HyperNeRF, while 4DGaussian takes about 20 minutes. In contrast, ADC-GS takes approximately 70 minutes. Although this additional training time is a limitation, the process remains efficient.
\noindent\textbf{Coding time}. The encoding/decoding process takes approximately 1.27 seconds
and 0.83 seconds under \(\lambda_e = e-2\), illustrating the practicality of the proposed method for real-world applications. 

\begin{table*}[t]
\centering
\caption{Per scene evaluation with PSNR and Size (MB) on HyperNeRF.}
\label{hypernerftable}
\setlength{\tabcolsep}{5pt} 
\renewcommand{\arraystretch}{1.2} 

\begin{tabular}{lcccccccccccc}
\toprule
\multirow{2}{*}{Method} & \multicolumn{2}{c}{3D Printer} & \multicolumn{2}{c}{Chicken} & \multicolumn{2}{c}{Broom} & \multicolumn{2}{c}{Banana} \\ 
                        & PSNR          & Size   & PSNR           & Size    & PSNR           & Size    & PSNR           & Size       \\ \midrule
\textbf{TiNeuVox-B}     & 22.80         & 48.00           & 28.30         & 48.00           & 21.50         & 48.00           & 24.40         & 48.00           \\ 
\textbf{FFDNeRF}        & 22.80         & 440.00          & 28.00         & 440.00          & 21.90         & 440.00          & 24.30         & 440.00          \\ 
\textbf{4DGaussian}           & 22.12         & 63.50           & 29.03         & 90.57           & 22.14         & 46.70           & 29.06         & 51.73           \\ 
\textbf{E-D3DGS}        & 22.77         & 28.50           & 29.83         & 51.00           & 21.70         & 31.60           & 28.98         & 73.20           \\ 
\midrule
\multirow{3}{*}{\textbf{Ours}} & 22.92         & 3.23            & 29.61         & 3.85            & 21.80         & 3.27            & 27.35         & 5.73            \\ 
                        & 23.01         & 4.05            & 29.69         & 5.12            & 21.89         & 4.73            & 27.53         & 6.90            \\ 
                        & 23.10         & 4.57            & 29.75         & 8.33            & 21.98         & 5.20            & 27.89         & 8.56            \\ 
\bottomrule
\end{tabular}
\end{table*}

\begin{table*}[t]
\centering
\caption{Per scene evaluation with PSNR and Size (MB) on Neu3D.}
\label{neu3dtable}
\setlength{\tabcolsep}{3.5pt} 
\renewcommand{\arraystretch}{1.2} 
\begin{tabular}{lcccccccccccccc}
\toprule
\multirow{2}{*}{Method} & \multicolumn{2}{c}{Cook Spinach} & \multicolumn{2}{c}{Cut Roasted Beef} & \multicolumn{2}{c}{Flame Salmon} & \multicolumn{2}{c}{Flame Steak} & \multicolumn{2}{c}{Sear Steak} \\ 
                        & PSNR          & Size     & PSNR           & Size     & PSNR           & Size  & PSNR           & Size  & PSNR           & Size  \\ \midrule
\textbf{HexPlane}       & 31.86         & 312.17       & 32.71          & 312.17       & 29.26          & 312.17     & 32.09         & 312.17    & 31.92          & 312.17     \\ 
\textbf{KPlanes}        & 32.60         & 579.7        & 31.82          & 579.7        & 30.44          & 579.7        & 32.39         & 579.7     & 32.52          & 579.7      \\ 
\textbf{MixVoxels}      & 31.65         & 500          & 31.30          & 500          & 29.92          & 500        & 31.21         & 500       & 31.43          & 500        \\ 
\textbf{4DGaussian}           & 32.41         & 38.28        & 32.37          & 38.80        & 29.26          & 38.46      & 31.86         & 36.16     & 32.70          & 36.16      \\ 
\textbf{CompGS}         & 30.75         & 824          & 29.90          & 810          & 25.20          & 864        & 30.96         & 825       & 31.23          & 822        \\ 
\textbf{E-D3DGS}        & 32.50         & 39.04        & 29.70          & 39.30        & 29.46          & 68.30      & 31.71         & 36.40     & 32.74          & 28.59      \\ 
\midrule
\multirow{2}{*}{\textbf{Ours}} & 32.02         & 4.07         & 31.61          & 3.85         & 28.65          & 4.41       & 32.48         & 3.80      & 32.29          & 4.08       \\ 
                              & 32.14         & 5.29         & 31.69          & 5.21         & 28.92          & 5.83       & 32.58         & 5.51      & 32.40          & 4.78       \\ 
                              & 32.34         & 6.77         & 31.88          & 6.48         & 29.01          & 6.08       & 32.65         & 7.59      & 32.48          & 5.95       \\ 
\midrule
\bottomrule
\end{tabular}
\end{table*}

\begin{figure}[t]
    \centering
    \includegraphics[width=\linewidth]{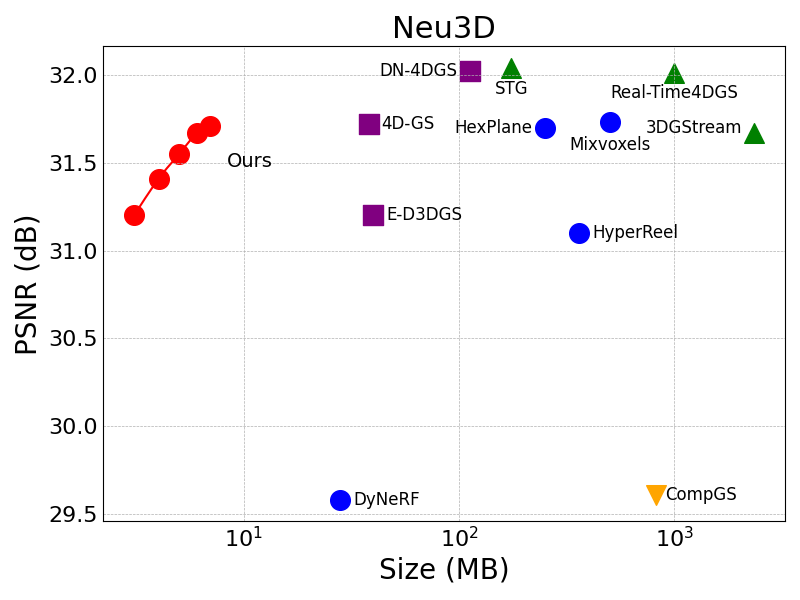}
    \caption{Rate-distortion curves of ADC-GS and comparison methods on Neu3D.}
    \label{rd-n3d}
\end{figure}
\section{Per-scene Results on Evaluation Datasets}
We present the per-scene evaluation results of ADC-GS compared with other methods. As shown in Tables \ref{hypernerftable} and \ref{neu3dtable}, we evaluate ADC-GS on two widely used datasets: HyperNeRF and Neu3D. HyperNeRF includes scenes such as 3D Printer, Chicken, Broom, and Banana, while Neu3D covers Cook Spinach, Cut Roasted Beef, Flame Salmon, Flame Steak, and Sear Steak. ADC-GS consistently achieves SOTA size efficiency with comparable rendering quality across all scenes. Notably, in some cases, such as 3D Printer and Flame Steak, ADC-GS also achieves the best rendering quality, demonstrating its superiority.

\section{Rate-Distortion performance on Neu3D.}

Figure \ref{rd-n3d} illustrates the R-D performance of ADC-GS in comparison to other methods on the Neu3D dataset. We set \(\lambda_e\) to {0.01, 0.005, 0.001, 0.0005, 0.0001} for five different bitrates. ADC-GS achieves up to 204\(\times\) size reduction without compromising rendering quality.

\section{Qualitative results.}
\begin{figure*}[t]
    \centering
    \includegraphics[width=\textwidth]{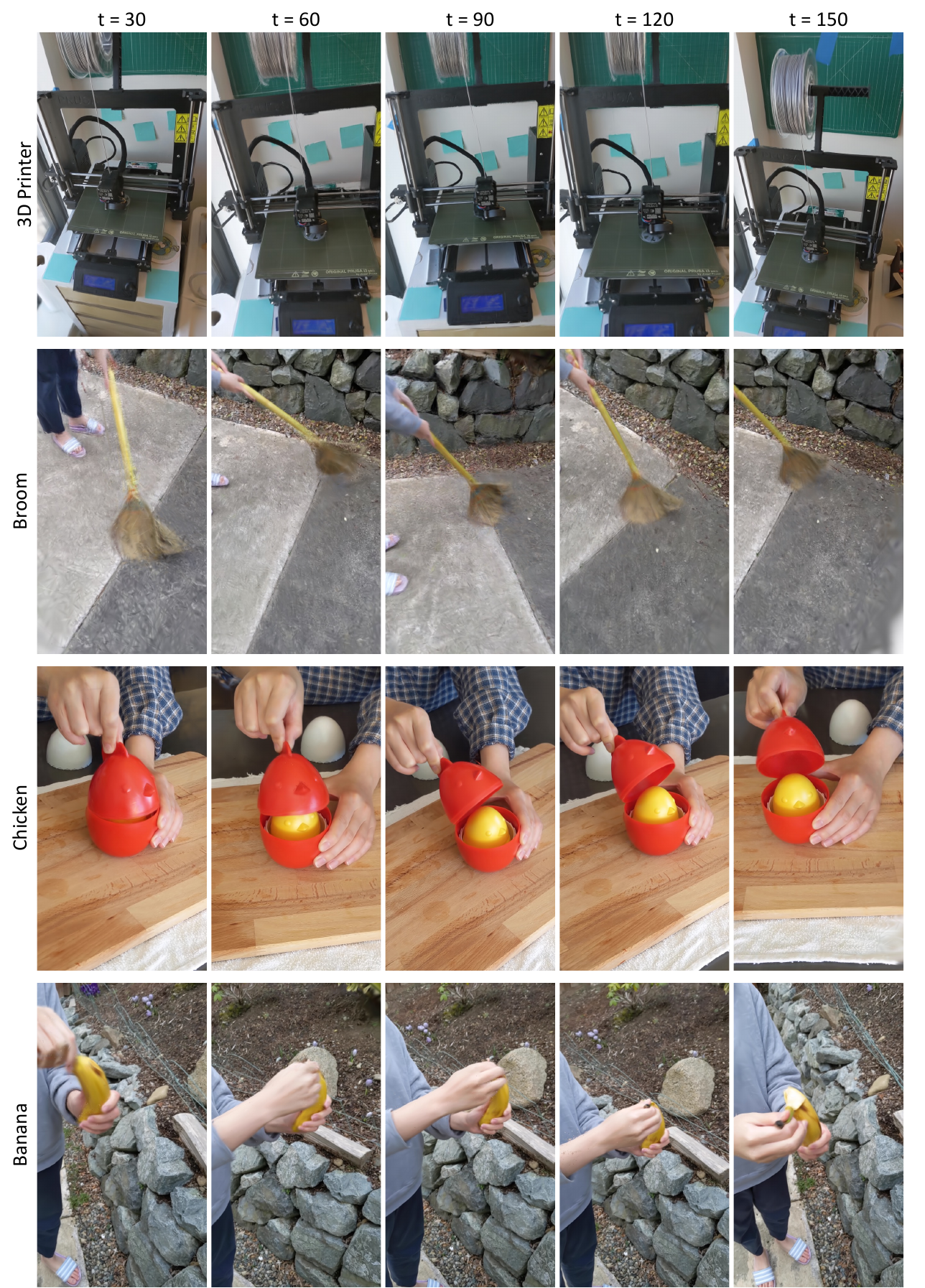}
    \vspace{-21pt}
    \caption{Qualitative results of HyperNeRF dataset.}
    \vspace{-10pt}
    \label{visual_hyper}
\end{figure*}

\begin{figure*}[t]
    \centering
    \includegraphics[width=\textwidth]{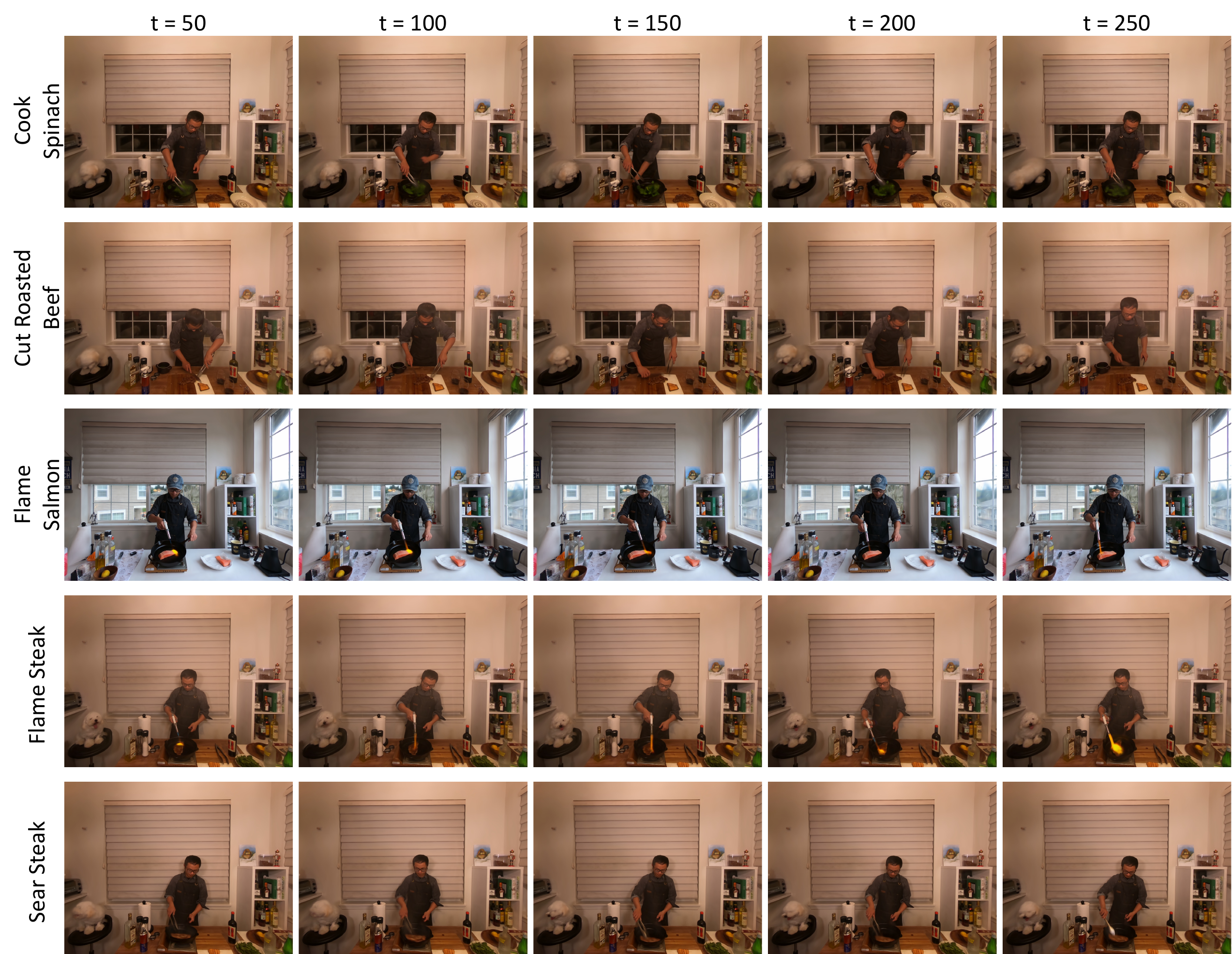}
    \vspace{-21pt}
    \caption{Qualitative results of Neu3D dataset.}
    \vspace{-10pt}
    \label{visual_neu}
\end{figure*}

Figures \ref{visual_hyper} and \ref{visual_neu} present additional visual results for the HyperNeRF and Neu3D datasets with \(\lambda_e = e-2\) across different frames.

\end{document}